\documentclass[10pt,twocolumn,letterpaper]{article}

\usepackage{iccv}
\usepackage{times}
\usepackage{epsfig}
\usepackage{graphicx}
\usepackage{amsmath}
\usepackage{amssymb}
\usepackage{multirow}
\usepackage{booktabs}
\usepackage{overpic}
\usepackage{dashrule}
\usepackage{blindtext, rotating}
\usepackage{color, colortbl}
\usepackage[misc]{ifsym}
\usepackage{subcaption}
\definecolor{Gray}{gray}{0.9}

\usepackage[pagebackref=true,breaklinks=true,letterpaper=true,colorlinks,bookmarks=false]{hyperref}

\iccvfinalcopy 

\def\iccvPaperID{2325} 

\newcommand\blfootnote[1]{%
  \begingroup
  \renewcommand\thefootnote{}\footnote{#1}%
  \addtocounter{footnote}{-1}%
  \endgroup
}

\begin{document}

\title{SparseBEV: High-Performance Sparse 3D Object Detection \\ from Multi-Camera Videos}

\author{
Haisong Liu $^{1}$ \quad Yao Teng $^{1}$ \quad Tao Lu $^{1}$ \quad Haiguang Wang $^{1}$ \quad Limin Wang $^{1,2,}$ \textsuperscript{\Letter} \\
$^{1}$State Key Laboratory for Novel Software Technology, Nanjing University \quad $^{2}$Shanghai AI Lab \\
\small\texttt{\{liuhs, yaoteng, taolu, haiguangwang\}@smail.nju.edu.cn, lmwang@nju.edu.cn}
}

\maketitle

\begin{abstract}
  Camera-based 3D object detection in BEV (Bird's Eye View) space has drawn great attention over the past few years.
  Dense detectors typically follow a two-stage pipeline by first constructing a dense BEV feature and then performing object detection in BEV space, which suffers from complex view transformations and high computation cost.
  On the other side, sparse detectors follow a query-based paradigm without explicit dense BEV feature construction, but achieve worse performance than the dense counterparts.
  In this paper, we find that the key to mitigate this performance gap is the adaptability of the detector in both BEV and image space.
  To achieve this goal, we propose SparseBEV, a fully sparse 3D object detector that outperforms the dense counterparts.
  SparseBEV contains three key designs, which are (1) scale-adaptive self attention to aggregate features with adaptive receptive field in BEV space, (2) adaptive spatio-temporal sampling to generate sampling locations under the guidance of queries, and (3) adaptive mixing to decode the sampled features with dynamic weights from the queries.
  On the test split of nuScenes, SparseBEV achieves the state-of-the-art performance of 67.5 NDS. On the val split, SparseBEV achieves 55.8 NDS while maintaining a real-time inference speed of 23.5 FPS.
  Code is available at \url{https://github.com/MCG-NJU/SparseBEV}.
\end{abstract}
\blfootnote{\Letter: Corresponding author.}

\section{Introduction}

Camera-based 3D Object Detection \cite{bevdet, detr3d, petr, bevdet4d, bevdepth, bevformer, solofusion} has witnessed great progress over the past few years. Compared with the LiDAR-based counterparts \cite{pointpillars, centerpoint, largekernel3d, link}, camera-based approaches have lower deployment cost and can detect long-range objects.

\begin{figure}[t]
  \includegraphics[width=0.95\linewidth]{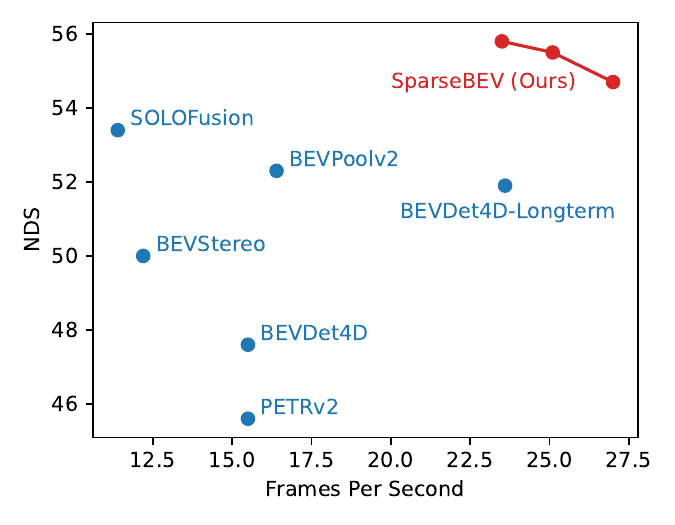}
  \vspace{-5pt}
  \caption{Performance comparison on the \texttt{val} split of nuScenes \cite{nuscenes}. All methods use ResNet50 \cite{resnet} as the image backbone and the input size is set to 704 $\times$ 256. FPS is measured on a single RTX 3090 with the PyTorch \texttt{fp32} backend. We balance accuracy and speed by reducing the number of decoder layers \textit{without} re-training.}
  \vspace{-10pt}
  \label{fig:intro}
\end{figure}

Previous methods can be divided into two paradigms. BEV (Bird's Eye View)-based methods \cite{bevdet, bevdet4d, bevformer, bevdepth, solofusion} follow a two-stage pipeline by first constructing an explicit \textit{dense} BEV feature from multi-view features and then performing object detection in BEV space. Although achieving remarkable progress, they suffer from high computation cost and rely on complex view transformation operators.
Another line of work \cite{detr3d, petr, petrv2} explores the \textit{sparse} query-based paradigm by initializing a set of sparse reference points in 3D space. Specifically, DETR3D \cite{detr3d} links the queries to image features using 3D-to-2D projection. It has simpler structure and faster speed, but its performance still lags far behind the dense ones. PETR series \cite{petr, petrv2} uses \textit{dense} global attention for the interaction between query and image feature, which is computationally expensive and buries the advantage of the sparse paradigm.
Thus, a natural question arises {\em whether {\bf fully sparse} detectors achieve similar accuracy to the dense ones?}

In this paper, we find that the key to obtain high performance in sparse 3D object detection is the \textit{adaptability} of the detector in both BEV and image space.
\textit{In BEV space}, the detector should be able to aggregate multi-scale features adaptively.
Dense BEV-based detectors typically use a BEV encoder to encode multi-scale BEV features. It can be a stack of residual blocks with FPN \cite{fpn} (e.g. BEVDet \cite{fpn}), or a transformer encoder with multi-scale deformable attention \cite{deformabledetr} (e.g. BEVFormer \cite{bevformer}).
For sparse detectors such as DETR3D, we argue that the multi-head self attention (MHSA) \cite{transformer} among queries can play the role of the BEV encoder, as queries are defined in BEV space.
However, the vanilla MHSA has a global receptive field, lacking an explicit multi-scale design.
\textit{In image space}, the detector should be adaptive to different objects with different sizes and categories.
This is because although the objects have similar sizes in 3D space, they might vary greatly in images.
However, the single-point sampling in DETR3D has a fixed local receptive field and the sampled feature is processed by static linear layers, hindering its performance.

To this end, we present SparseBEV, a \textit{fully sparse} 3D object detector that matches or even outperforms the dense counterparts.
Our SparseBEV detector contains three key designs, which are (1) \textit{scale-adaptive self attention} to aggregate features with adaptive receptive field in BEV space, (2) \textit{adaptive spatio-temporal sampling} to generate sampling locations under the guidance of queries, and (3) \textit{adaptive mixing} to decode the sampled features with dynamic weights from the queries.
We also propose to use \textit{pillars} instead of reference points as the formulation of query, since pillars introduce better spatial priors.

We conduct comprehensive experiments on the nuScenes dataset. As shown in Fig. \ref{fig:intro}, our SparseBEV achieves the performance of 55.8 NDS and the speed of 23.5 FPS on the \texttt{val} split, surpassing all previous methods in both speed and accuracy. Besides, we can flexibly adjust of the inference speed by reducing the number of decoder layers without re-training. On \texttt{test} split, SparseBEV with V2-99 \cite{vovnet2} backbone achieves 63.6 NDS \textit{without} using future frames or test-time augmentation. By further utilizing future frames, SparseBEV achieves 67.5 NDS, outperforming the previous state-of-the-art BEVFormerV2 \cite{bevformerv2} by 2.7 NDS.

\section{Related Work}

\subsection{Query-based 2D Object Detection}

Recently, Transformer~\cite{transformer} with its attention blocks has been widely applied in the computer vision tasks~\cite{vit,segnext,detr}. In object detection, DETR~\cite{detr} was the first model to predict objects based on learnable queries and treat the detection as a set prediction problem. A lot of works~\cite{deformabledetr,sparsercnn,dabdetr,conditionaldetr,adamixer,dndetr,dinodetr} were then proposed to accelerate the convergence of DETR by using the sampled features instead of using the global ones. For example, Deformable-DETR~\cite{deformabledetr} samples image features based on sparse reference points, and then applies the deformable attention on the features. Sparse R-CNN~\cite{sparsercnn} uses the ROIAlign~\cite{maskrcnn} to obtain local features and then performs the dynamic convolution. AdaMixer~\cite{adamixer} combines the advantages of the sampling points and the dynamic convolution to further boost the convergence. DN-DETR~\cite{dndetr} devises a denoising mechanism that feeds ground-truth bounding boxes with noises into decoder and trains the model to reconstruct the original boxes. DINO~\cite{dinodetr} furtherly optimizes DN-DETR~\cite{dndetr} and DAB-DETR~\cite{dabdetr} by proposing a contrastive denoising training strategy and a mixed query selection method. There are also several methods~\cite{groupdetr, hybriddetr, stageinteractor} designed to tackle the instability of the training procedure. Our work follows the query-based detection paradigm and extends it to 3D space with temporal information.

\subsection{Monocular 3D Object Detection}

Monocular 3D object detection takes one single image as input and outputs predicted 3D bounding boxes of objects. A significant challenge in monocular 3D object detection is how to transfer 2D features to 3D space. Several works~\cite{pseudo-lidar, caddn, fcos3d, dd3d} incorporates depth information to deal with this problem. 
Pseudo LiDAR~\cite{pseudo-lidar} first estimates the depth of input images and constructs pseudo point clouds. The pseudo point clouds are then sent to a LiDAR-based detection module to predict the 3D boxes of interested objects. 
CaDDN~\cite{caddn} further proposes a fully differentiable end-to-end network which learns pixel-wise categorical depth distributions to predict appropriate depth intervals in 3D space. 
Inspired by FCOS \cite{fcos}, FCOS3D~\cite{fcos3d} projects 3D coordinates onto 2D images and decouples them as 2D attributes (centerness and classification) and 3D attributes (depths, sizes, and orientations).

\subsection{3D Object Detection in BEV}

\begin{figure*} 
  \centering
  \includegraphics[width=\linewidth]{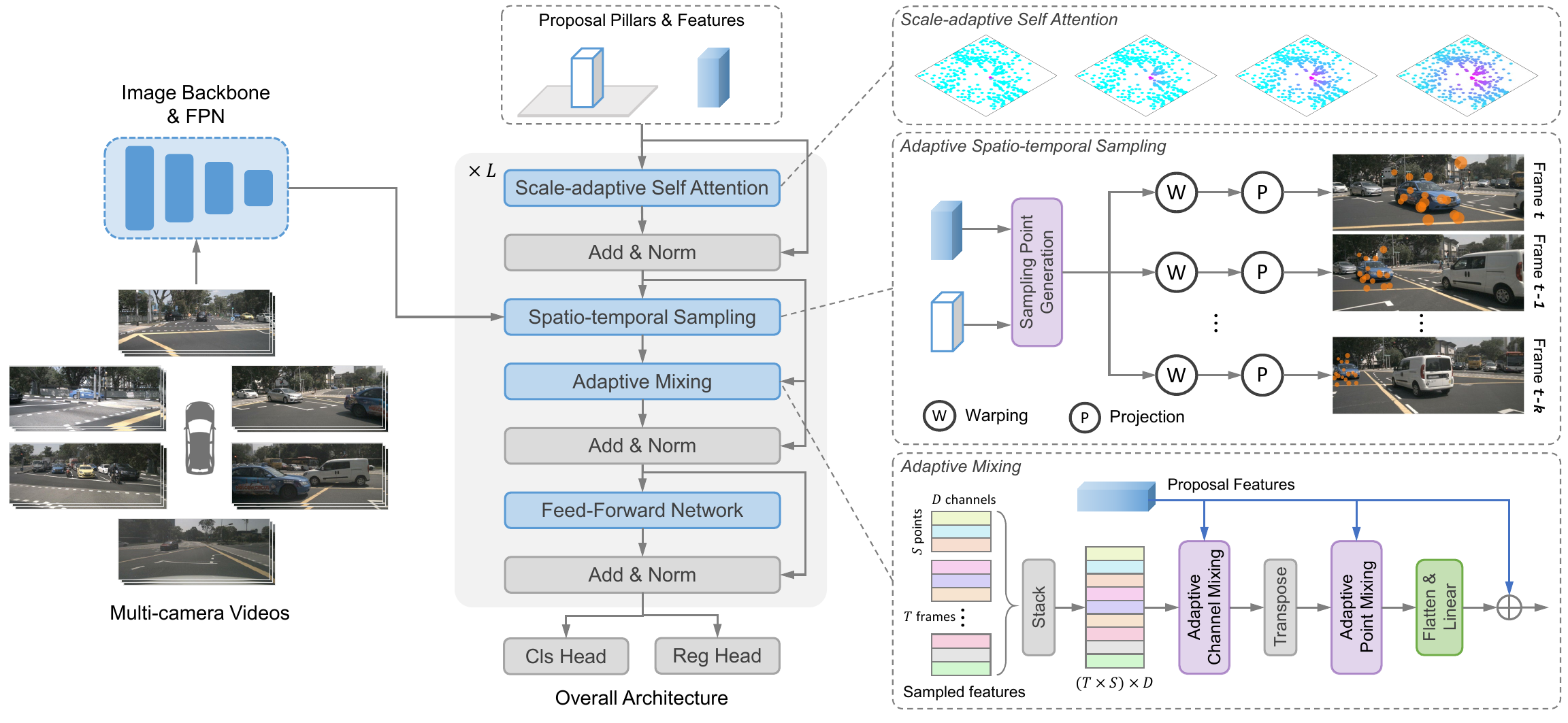}
  \vspace{-12pt}
  \caption{The overall architecture of SparseBEV, a fully-sparse camera-only 3D object detector. Queries are initialized to be a sparse set of pillars in BEV space. The scale-adaptive self attention further encodes the queries with adaptive receptive fields. Next, multi-view and multi-timestamp features are aggregated with adaptive spatio-temporal sampling and decoded by adaptive mixing. The decoder repeats $L$ times to produce final predictions.}
  \vspace{-5pt}
  \label{fig:arch}
\end{figure*}

Bird's-eye-view (BEV) object detection \cite{vpn, lss, bevdet, detr3d, petr, simmod, bevformer, bevformerv2, solofusion} aims at detecting objects in BEV space given either single-view or multi-view 2D images.
Early works~\cite{oft, pseudo-lidar, caddn} typically transformed 2D features to BEV space based on single-view images and conducted monocular 3D object detection.
LSS~\cite{lss} takes six 2D images of different views as input and transforms them into 3D space based on depth estimation.
Based on LSS, BEVDet~\cite{bevdet} lifts 2D feature to BEV space and uses a BEV encoder with residual blocks and FPN to further encode the BEV features.
BEVFormer~\cite{bevformer} proposes a spatio-temporal transformer encoder that projects multi-view and multi-timestamp input to BEV representations.
To ease the optimization, BEVFormerV2~\cite{bevformerv2} introduces perspective view supervision and supervises monocular 3D object detection parallel with BEV object detection.
SOLOFusion~\cite{solofusion} defines the localization potential for depth estimation and fuses short-term, high-resolution and long-term, low-resolution temporal stereo.

Inspired by DETR, another line of works \cite{detr3d, petr, petrv2} explore the sparse query-based paradigm.
DETR3D~\cite{detr3d} proposes a top-down framework starting from a learnable sparse set of reference points and refining them iteratively via 3D-to-2D queries.
However, such 3D-to-2D projection hinders the receptive field of the query.
To handle this, PETR series \cite{petr, petrv2} use global attention for the interaction betweeen queries and image features, and introduce 3D positional embeddings to encode 2D features into 3D representation without explicit projection.
Although achieving notable improvements, the expensive dense global attention buries the advantages of the sparse paradigm and makes it difficult to utilize long-term temporal information efficiently.
In contrast, we keep the fully sparse design of DETR3D and boost the performance by enhancing the adaptability of the detector.

\section{SparseBEV}

As shown in Fig. \ref{fig:arch}, SparseBEV is a query-based one-stage detector with $L$ decoder layers. The input multi-camera videos are processed frame-by-frame using image backbone and FPN. Next, a set of sparse pillar queries are initialized in BEV space and aggregated by scale-adaptive self attention. These queries then interact with the image features via adaptive spatio-temporal sampling and adaptive mixing to make 3D object predictions. We also propose a dual-branch version of SparseBEV to further enhance the temporal modeling.

\subsection{Query Formulation}

We first define a set of learnable queries, where each of them is represented by its translation $[x, y, z]$, dimension $[w, l, h]$, rotation $\theta$ and velocity $[v_x, v_y]$. The queries are initialized to be \textit{pillars} in BEV space where $z$ is set to 0 and $h$ is set to $\sim$4m. The initial velocity is set to $[v_x, v_y] = \mathbf{0}$. Other parameters ($x, y, w, l, \theta$) are drawn from random gaussian distributions. Following Sparse R-CNN\cite{sparsercnn}, we attach a $D$-dim query feature to each query box to encode the rich instance characteristics.

\subsection{Scale-adaptive Self Attention}

As mentioned above, dense BEV-based methods typically use a BEV encoder to encode multi-scale BEV features. However, in our method, since we do not explicitly build a BEV feature, how to aggregate multi-scale features in BEV space is a challenge.

In this work, we argue that the self attention can play the role of BEV encoder, since queries are defined in BEV space. The vanilla multi-head self attention has global receptive field and lacks the ability of local multi-scale aggregation. Thus, we propose \textit{scale-adaptive self attention} (SASA), which learns appropriate receptive fields under the guidance of queries. First, we compute the all-pair distances $D \in \mathbb{R}^{N \times N}$ ($N$ is the number of queries) between the query centers in BEV space:
\begin{equation}
  D_{i, j} = \sqrt{(x_i - x_j)^2 + (y_i - y_j)^2},
\end{equation}
where $x_i$ and $y_i$ denotes the center of the $i$-th query. The attention considers not only the similarity between query features, but the distance between them as well. A toy example below shows how it works:
\begin{align}
  \text{Attn}(Q, K, V) &= \text{Softmax}(\frac{QK^T}{\sqrt{d}} - \tau D) V,
\end{align}
where $Q, K, V \in \mathbb{R}^{N \times d}$ is the query itself and $d$ is the channel dimension. $\tau$ is a scalar to control the receptive field for each query. When $\tau = 0$, it degrades to the vanilla self attention with global receptive field. As $\tau$ grows, the attention weights for distant queries becomes smaller and the receptive field narrows.

In practice, the receptive field controller $\tau$ is adaptive to each query and specific to each head. Supposing there are $H$ heads, we use a linear transformation to generate head-specific $\tau_1, \tau_2, ..., \tau_H$ from the give query $\mathbf{q} \in \mathbb{R}^d$:
\begin{equation}
  [\tau_1, \tau_2, ..., \tau_H] = \text{Linear}_{d \to H} (\mathbf{q}) \in \mathbb{R}^{H},
\end{equation}
where the weights are shared across different queries.

\begin{figure}[t]
  \centering
  \includegraphics[width=\linewidth]{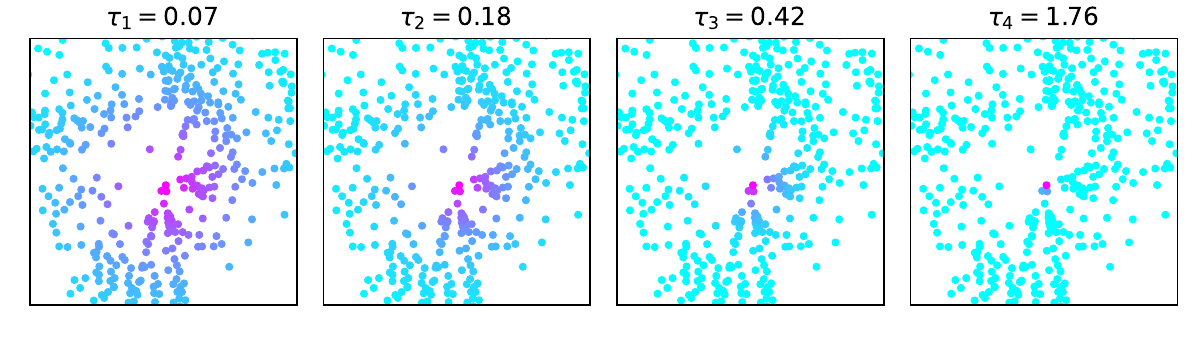}
  \includegraphics[width=\linewidth]{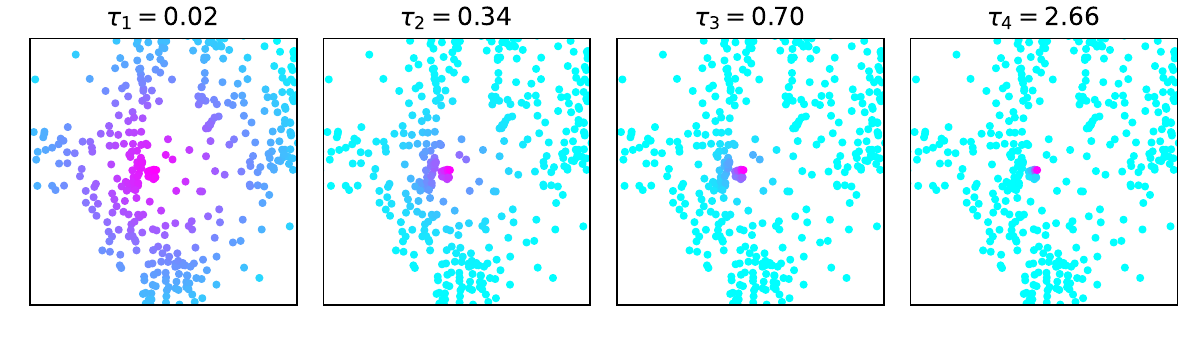}
  \vspace{-20pt}
  \caption{Visualization of the learned BEV receptive field in SASA. We choose two samples from the validation split and visualize 4 heads for each sample, denoted by $\tau_1$, $\tau_2$, $\tau_3$ and $\tau_4$. Queries are represented by their centers.}
  \label{fig:mssa_receptive_field}
  \vspace{-5pt}
\end{figure}

In our experiments, we surprisingly find that the $\tau$ for each head is learnt to uniformly distribute in a certain range regardless of the initialization. In Fig. \ref{fig:mssa_receptive_field}, we sort the heads according to $\tau$ and visualize the attention weights for the distance part. As we can see, each head learns a different receptive field from each other, enabling the network to aggregate features in a multi-scale manner like FPN.

Scale-adaptive self attention (SASA) demonstrates the necessity of FPN, while being more flexible as it learns the scales adaptively from the query. We also find an interesting phenomenon that different categories of queries have different sizes of receptive field. For example, queries representing the bus have larger receptive field than those representing the pedestrians. More details can be found in the ablation studies.

\subsection{Adaptive Spatio-temporal Sampling}

For each frame, we use a linear layer to generate a set of sampling offsets $\{(\Delta x_i, \Delta y_i, \Delta z_i)\}$ adaptively from the query feature. These offsets are transformed to 3D sampling points based on the query pillar:

\begin{align}
  \left[\!\begin{array}{c}
    x_i \\
    y_i \\
    z_i \\
  \end{array}\!\right] \! = \! 
  \left[\!\!\begin{array}{cccc}
    \cos\theta\!\!\! & -\!\sin\theta\!\!\! & 0\! \\
    \sin\theta\!\!\! & \cos\theta\!\!\! & 0\! \\
    0\!\!\! & 0\!\!\! & 1\! \\
  \end{array}\!\!\right]\!\!
  \left[\!\!\begin{array}{c}
    w \!\cdot\! \Delta x_i \\
    l \!\cdot\! \Delta y_i \\
    h \!\cdot\! \Delta z_i \\
  \end{array}\!\!\right] \! + \!
  \left[\!\begin{array}{c}
    x \\
    y \\
    z \\
  \end{array}\!\right].
\end{align}

Compared with the deformable attention in BEVFormer, our sampling points are adaptive to both query pillar and query feature, thus better covering objects with varying sizes. Besides, these points are not restricted to the query, since we do not limit the range of the sampling offsets.

Next, we perform temporal alignment by warping the sampling points according to motions. In autonomous driving, there are two types of motion: one is ego-motion and the other is object motion. Ego-motion describes the motion of the car from its own perspective as it navigates through the environment, while object motion refers to the movement of other objects in the environment as they move around the self-driving car.

\vspace{-5pt}
\paragraph{Dealing with Object Motion.} As mentioned above, instantaneous velocity can be equal to average velocity for a short time window in self-driving. Thus, we adaptively warp the sampling points to previous timestamps using the velocity vector $[v_x, v_y]$ from the query:
\begin{align}
  x_{t, i} &= x_i + v_x \cdot (T_t - T_0) \\
  y_{t, i} &= y_i + v_y \cdot (T_t - T_0),
\end{align}
where $T_t$ denotes the timestamp of previous frame $t$ ($T_0$ denotes the current timestamp). Note that $z_{t, i}$ is identical to $z_i$ because the velocity vector is defined in BEV plane.

\vspace{-5pt}
\paragraph{Dealing with Ego Motion.} Next, we warp the sampling points based on the ego pose provided by the dataset. Points are first transformed to the global coordinate system and then to the local coordinate system of frame $t$:
\begin{equation}
  [x_{t, i}' \ \ \ y_{t, i}' \ \ \ z_{t, i}' \ \ \ 1]^T = E_t E_0^{-1}
  [x_{t, i} \ \  y_{t, i} \ \ z_{t, i} \ \ 1]^T,
\end{equation}
where $E_t = [\mathbf{R} | \mathbf{t}]$ is the ego pose of frame $t$.

\setlength{\tabcolsep}{4pt}
\begin{table*}[t]
   \centering
   \begin{tabular}{l|ccc|cc|ccccc}
      \toprule
      Method & Backbone & Input Size & Epochs & NDS & mAP & mATE & mASE & mAOE & mAVE & mAAE \\
      \midrule
      PETRv2 \cite{petrv2}         & ResNet50 & 704 $\times$ 256 & 60            & 45.6 & 34.9 & 0.700 & 0.275 & 0.580 & 0.437 & 0.187 \\
      BEVStereo \cite{bevstereo}   & ResNet50 & 704 $\times$ 256 & 90 $\ddagger$ & 50.0 & 37.2 & 0.598 & 0.270 & 0.438 & 0.367 & 0.190 \\
      BEVPoolv2 \cite{bevpoolv2}   & ResNet50 & 704 $\times$ 256 & 90 $\ddagger$ & 52.6 & 40.6 & 0.572 & 0.275 & 0.463 & 0.275 & 0.188 \\
      SOLOFusion \cite{solofusion} & ResNet50 & 704 $\times$ 256 & 90 $\ddagger$ & 53.4 & 42.7 & 0.567 & 0.274 & 0.511 & 0.252 & 0.181 \\
      Sparse4Dv2 \cite{sparse4dv2} & ResNet50 & 704 $\times$ 256 & 100 & 53.9 & 43.9 & 0.598 & 0.270 & 0.475 & 0.282 & 0.179 \\
      StreamPETR $\dagger$ \cite{streampetr} & ResNet50 & 704 $\times$ 256 & 60 & 55.0 & \textbf{45.0} & 0.613 & 0.267 & 0.413 & 0.265 & 0.196  \\
      \rowcolor{Gray}
      \textbf{SparseBEV}           & ResNet50 & 704 $\times$ 256 & 36 & 54.5 & 43.2 & 0.606 & 0.274 & 0.387 & 0.251 & 0.186 \\
      \rowcolor{Gray}
      \textbf{SparseBEV} $\dagger$ & ResNet50 & 704 $\times$ 256 & 36 & \textbf{55.8} & 44.8 & 0.581 & 0.271 & 0.373 & 0.247 & 0.190 \\
      \midrule
      DETR3D $\dagger$ \cite{detr3d}       & ResNet101-DCN & 1600 $\times$ 900 & 24 & 43.4 & 34.9 & 0.716 & 0.268 & 0.379 & 0.842 & 0.200 \\
      BEVFormer $\dagger$ \cite{bevformer} & ResNet101-DCN & 1600 $\times$ 900 & 24 & 51.7 & 41.6 & 0.673 & 0.274 & 0.372 & 0.394 & 0.198 \\
      BEVDepth \cite{bevdepth}             & ResNet101 & 1408 $\times$ 512 & 90 $\ddagger$ & 53.5 & 41.2 & 0.565 & 0.266 & 0.358 & 0.331 & 0.190 \\
      Sparse4D $\dagger$ \cite{sparse4d}   & ResNet101-DCN & 1600 $\times$ 900 & 48 & 55.0 & 44.4 & 0.603 & 0.276 & 0.360 & 0.309 & 0.178 \\
      SOLOFusion \cite{solofusion}         & ResNet101 & 1408 $\times$ 512 & 90 $\ddagger$ & 58.2 & 48.3 & 0.503 & 0.264 & 0.381 & 0.246 & 0.207 \\
      \rowcolor{Gray}
      \textbf{SparseBEV} $\dagger$         & ResNet101 & 1408 $\times$ 512 & 24 & \textbf{59.2} & \textbf{50.1} & 0.562 & 0.265 & 0.321 & 0.243 & 0.195 \\
      \bottomrule
   \end{tabular}
   \vspace{-5pt}
   \caption{Performance comparison on the nuScenes \texttt{val} split. $\dagger$ benefits from perspective pretraining. $\ddagger$ indicates methods with CBGS \cite{cbgs} which will elongate 1 epoch into 4.5 epochs.}
   \label{table:nuscenes_val}
\end{table*}

\vspace{-5pt}
\paragraph{Sampling.} For each timestamp, we project the warped sampling points $\{(x_{t, i}', y_{t, i}', z_{t, i}')\}$ to each view using the provided camera intrinsics and extrinsics. Since there are overlaps between adjacent views, the projected point may hit one or more views, which are termed as $\mathcal{V}$. For each hit view $k$, we have multi-scale feature maps $\{F_{k, j} | j \in \{1, 2, ... ,N_{\text{feat}}\} \}$ from the image backbone. Features are first sampled by bilinear interpolation $\mathbf{B}$ in the image plane and then weighted over the scale axis:
\begin{equation}
  f_{t, i} = \frac{1}{|\mathcal{V}|} \sum_{k \in \mathcal{V}} \sum_{j=1}^{N_{\text{feat}}} w_{ij} \mathbf{B} (F_{k, j}, \mathbf{P}_k(x_{t, i}', y_{t, i}', z_{t, i}'))
\end{equation}
where $N_{\text{feat}}$ is the number of the multi-scale feature maps and $\mathbf{P}_k$ is the projection function for view $k$. $w_{ij}$ is the weight for the $i$-th point on the $j$-th feature map and is generated from the query feature by linear transformation.

\subsection{Adaptive Mixing}

Given the sampled features from different timestamps and locations, the key is how to adaptively decode it under the guidance of queries. Inspired by AdaMixer \cite{adamixer}, we introduce a simple but effective approach to decode and aggregate the spatio-temporal features with dynamic convolutions \cite{dynamic_conv} and MLP-Mixer \cite{mlpmixer}.
Supposing there are $T$ frames in total and $S$ sampling points per frame, we first stack them to a total of $P=T \times S$ points. Thus, the sampled features are organized to $f \in \mathbb{R}^{P \times C}$.

\paragraph{Channel Mixing.} We first perform channel mixing on $f$ to enhance object semantics. The dynamic weights are generated from the query feature $\mathbf{q}$:
\begin{align}
  W_c &= \text{Linear}(\mathbf{q}) \in \mathbb{R}^{C \times C} \\
  \text{M}_c(f) &= \text{ReLU}(\text{LayerNorm}(f W_c)),
\end{align}
where $W_c$ is the dynamic weights and is shared across different frames and different sampling points.

\paragraph{Point Mixing.} Next, we then transpose the feature and apply the dynamic weights to the point dimension of it:
\begin{align}
  W_p &= \text{Linear}(\mathbf{q}) \in \mathbb{R}^{P \times P} \\
  \text{M}_p(f) &= \text{ReLU}(\text{LayerNorm}(f^T W_p)),
\end{align}
where $W_p$ is the dynamic weights and is shared across different channels.

After channel mixing and point mixing, the spatio-temporal features are flattened and aggregated by a linear layer. The final regression and classification predictions are computed by two MLPs respectively.

\setlength{\tabcolsep}{4pt}
\begin{table*}[t]
  \centering
  \begin{tabular}{l|c|c|cc|ccccc}
    \toprule
    Method & Backbone & Epochs & NDS & mAP & mATE & mASE & mAOE & mAVE & mAAE \\ 
    \midrule
    DETR3D \cite{detr3d}           & V2-99 & 24 & 47.9 & 41.2 & 0.641 & 0.255 & 0.394 & 0.845 & 0.133 \\
    PETR \cite{petr}               & V2-99 & 24 & 50.4 & 44.1 & 0.593 & 0.249 & 0.383 & 0.808 & 0.132 \\
    UVTR \cite{uvtr}               & V2-99 & 24 & 55.1 & 47.2 & 0.577 & 0.253 & 0.391 & 0.508 & 0.123 \\
    BEVFormer \cite{bevformer}     & V2-99 & 24 & 56.9 & 48.1 & 0.582 & 0.256 & 0.375 & 0.378 & 0.126 \\
    BEVDet4D \cite{bevdet4d}       & Swin-B \cite{swin} & 90$^\ddagger$ & 56.9 & 45.1 & 0.511 & 0.241 & 0.386 & 0.301 & 0.121 \\
    PolarFormer \cite{polarformer} & V2-99 & 24 & 57.2 & 49.3 & 0.556 & 0.256 & 0.364 & 0.440 & 0.127 \\
    PETRv2 \cite{petrv2}           & V2-99 & 24 & 59.1 & 50.8 & 0.543 & 0.241 & 0.360 & 0.367 & 0.118 \\
    Sparse4D \cite{sparse4d}       & V2-99 & 48 & 59.5 & 51.1 & 0.533 & 0.263 & 0.369 & 0.317 & 0.124 \\
    BEVDepth \cite{bevdepth}       & V2-99 & 90$^\ddagger$ & 60.0 & 50.3 & 0.445 & 0.245 & 0.378 & 0.320 & 0.126 \\
    BEVStereo \cite{bevstereo}     & V2-99 & 90$^\ddagger$ & 61.0 & 52.5 & 0.431 & 0.246 & 0.358 & 0.357 & 0.138 \\
    SOLOFusion \cite{solofusion}   & ConvNeXt-B \cite{convnext} & 90$^\ddagger$ & 61.9 & 54.0 & 0.453 & 0.257 & 0.376 & 0.276 & 0.148 \\
    \rowcolor{Gray}
    \textbf{SparseBEV}             & V2-99 & 24 & \textbf{62.7} & \textbf{54.3} & 0.502 & 0.244 & 0.324 & 0.251 & 0.126 \\
    \rowcolor{Gray}
    \textbf{SparseBEV} (dual-branch) & V2-99 & 24 & \textbf{63.6} & \textbf{55.6} & 0.485 & 0.244 & 0.332 & 0.246 & 0.117 \\
    \midrule
    BEVFormerV2 $\dagger$ \cite{bevformerv2} & InternImage-XL \cite{internimage} & 24 & 64.8 & 58.0 & 0.448 & 0.262 & 0.342 & 0.238 & 0.128 \\
    BEVDet-Gamma $\dagger$ \cite{bevpoolv2}  & Swin-B & 90$^\ddagger$ & 66.4 & 58.6 & 0.375 & 0.243 & 0.377 & 0.174 & 0.123 \\
    \rowcolor{Gray}
    \textbf{SparseBEV} $\dagger$             & V2-99 & 24 & \textbf{67.5} & \textbf{60.3} & 0.425 & 0.239 & 0.311 & 0.172 & 0.116 \\
    \bottomrule
  \end{tabular}
  \vspace{-5pt}
  \caption{Performance comparison on the nuScenes \texttt{test} split. $\dagger$ uses future frames. $\ddagger$ indicates methods with CBGS \cite{cbgs} which will elongate 1 epoch into 4.5 epochs.}
  \label{table:nuscenes_test}
\end{table*}

\subsection{Dual-branch SparseBEV}

Inspired by SlowFast \cite{slowfast}, we further enhance the long-term temporal modeling by dividing the input video into a \textbf{slow} stream and a \textbf{fast} stream, resulting in a dual-branch SparseBEV.
The slow stream is designed to capture fine-grained appearance details and it operates at low frame rates and high resolutions. The fast stream is responsible for capturing long-term temporal stereo and it operates at high frame rates and low resolutions.
Sampling points are projected to the two streams respectively and the sampled features are stacked before adaptive mixing.

In this way, we decouple the learning of static appearance and temporal motion, leading to better performance. Besides, the computation cost is significantly reduced since only a small fraction of frames needs to be processed at high resolution. However, since this dual-branch design makes the framework a little complex, we do \textbf{not} use it unless otherwise stated. The ablations of this part can be found in the supplementary material.

\section{Experiments}

\subsection{Implementation Details}

We implement our model using PyTorch \cite{pytorch}. Following previous methods \cite{detr3d, bevformer, petrv2, solofusion}, we adopt common image backbones including ResNet \cite{resnet} and V2-99 \cite{vovnet2}.
The decoder consists of 6 layers and weights are shared across different layers. By default, we use $T=8$ frames in total and the interval between adjacent frames is about 0.5s.

During training, we adopt the Hungarian algorithm \cite{hungarian} for label assignment between ground-truth objects and predictions. Focal loss \cite{focalloss} is used for classification and L1 loss is used for 3D bounding box regression. We use the AdamW \cite{adamw} optimizer with a global batch size of 8. The initial learning rate is set to $2 \times 10^{-4}$ and is decayed with cosine annealing policy.

Recently, we follow the training setting of the very recent work StreamPETR \cite{streampetr} and refresh our results for fair comparison. For the regression loss, we change the loss weight of $x$ and $y$ to 2.0 and leave the others to 1.0. Query denoising \cite{dndetr} is also introduced to stablize training and speedup convergence.

\subsection{Datasets and Metrics}

We evaluate our model on the nuScenes dataset \cite{nuscenes}, which consists of large-scale multimodal data collected from 6 surround-view cameras, 1 lidar and 5 radars. The dataset has 1000 videos and is split into 700/150/150 videos for training/validation/testing. Each video has roughly 20s duration and the key samples are annotated every 0.5s.

For 3D object detection, there are up to 1.4M annotated 3D bounding boxes of 10 classes. The official evaluation metrics include the well-known mean Average Precision (mAP) and five true positive (TP) metrics, including ATE, ASE, AOE, AVE, and AAE for measuring translation, scale, orientation, velocity, and attribute errors respectively. The overall performance is measured by the nuScenes Detection Score (NDS), which is the composite of the above metrics.

\subsection{Comparison with the State-of-the-art Methods}

\paragraph{nuScenes \texttt{val} split.} In Tab. \ref{table:nuscenes_val}, we compare SparseBEV with previous state-of-the-art methods on the validation split of nuScenes.
Unless otherwise stated, the image backbone is pretrained on ImageNet-1k \cite{imagenet} and the number of queries is set to 900.
To keep the simplicity of our approach, the dual-branch design is not used here.
When adopting ResNet50 as the backbone and 704 $\times$ 256 as the input size, SparseBEV outperforms the previous state-of-the-art method SOLOFusion by 1.1 NDS and 0.5 mAP.
By further adopting nuImages \cite{nuscenes} pretraining and reducing the number queries to 400, SparseBEV sets a new record of 55.8 NDS while maintaining a real-time inference speed of 23.5 FPS (corresponding to Fig. \ref{fig:intro}).
Next, we upgrade the backbone to ResNet101 and scale the input size up to 1408 $\times$ 512.
Under this setting, SparseBEV still surpasses SOLOFusion by 1.8 mAP and 1.0 NDS, demonstrating the scalability of our method.

\vspace{-10pt}
\paragraph{nuScenes \texttt{test} split.} We submit our method to the website of nuScenes and report the leaderboard in Tab. \ref{table:nuscenes_test}.
Using the V2-99 \cite{vovnet2} backbone pretrained by DD3D \cite{dd3d}, SparseBEV achieves 62.7 NDS and 54.3 mAP without bells and whistles.
Built on top of this, the dual-branch design gives us extra bonus on the performance, hitting 63.6 NDS and 55.6 mAP.
We also follow the offline setting of BEVFormerV2 \cite{bevformerv2} which leverages both past and future frames to assist the detection.
Remarkably, our method (single branch only) surpasses BEVFormerV2 by 2.8 mAP and 2.2 NDS.
In addition, BEVFormerV2 uses InternImage-XL \cite{internimage} with over 300M parameters as the image backbone, while our V2-99 backbone is much more lightweight with only $\sim$70M parameters.

\subsection{Ablation Studies}

In this section, we conduct ablations on the validation split of nuScenes. For all experiments, we use ResNet50 pretrained on nuImages \cite{nuscenes} as the image backbone. The input contains 8 frames with 704 $\times$ 256 resolution. We use 900 queries in the decoder and the model is trained for 24 epochs without CBGS \cite{cbgs}.

\begin{table}[t]
  \centering
  \begin{tabular}{c|cc}
    \toprule
    Query Formulation & NDS & mAP \\
    \midrule
    3D Reference Points & 55.1 & 44.0 \\
    BEV Pillars & \textbf{55.6} & \textbf{45.4} \\
    \bottomrule
  \end{tabular}
  \caption{Ablations on query formulation. Compared with reference points, pillars introduce better spatial priors and lead to better performance.}
  \label{table:ablation_query}
\end{table}

\vspace{-5pt}
\paragraph{Query Formulation.} In Tab. \ref{table:ablation_query}, we ablate different formulations of the query. The top row uses a set of reference points distributing uniformly in 3D space (e.g. DETR3D and PETR). By replacing the reference points with pillars in BEV space, we observe performance improvements (+0.5 NDS) over the baseline. This is because pillars introduce better spatial priors than reference points.

\begin{table}[t]
  \centering
  \begin{tabular}{c|c|cc}
    \toprule
    Self Attention & Distance Function & NDS & mAP \\
    \midrule
    MHSA & - & 53.4 & 41.4 \\
    \midrule
    \multirow{3}{*}{SASA} & $\tau D^2$ & 54.3 & 43.8 \\
    & $\tau D$ & \textbf{55.6} & \textbf{45.4} \\
    & $\tau \sqrt{D}$ & 55.1 & 44.3 \\
    \bottomrule
  \end{tabular}
  \caption{Comparison between vanilla multi-head self attention (MHSA) and scale-adaptive self attention (SASA). Our SASA achieves significant improvements over the baseline.}
  \label{table:mssa}
  \vspace{-5pt}
\end{table}

\begin{figure}[t]
  \centering
  \vspace{-5pt}
  \subfloat{\includegraphics[width=\linewidth]{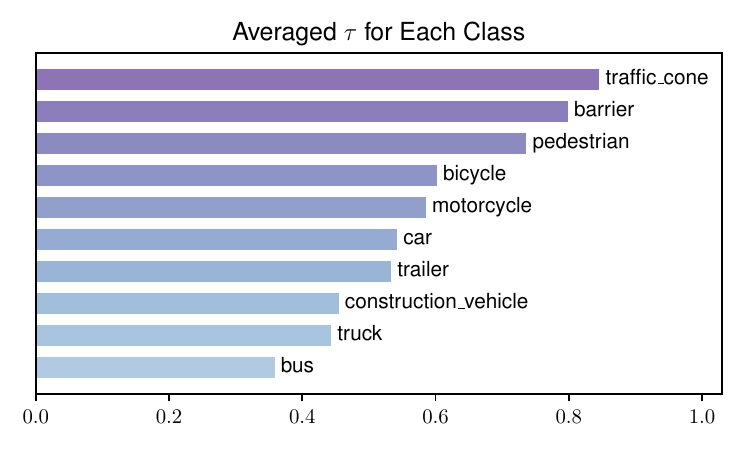}}%
  \vspace{-10pt}
  \caption{Averaged $\tau$ over all queries and all heads for each class in SASA. Larger $\tau$ indicates smaller receptive fields. We perform statistics on the val split of nuScenes and choose the queries with a confidence score over 0.3.}
  \label{fig:sasa_class}
\end{figure}

\vspace{-5pt}
\paragraph{Scale-adaptive Self Attention.} We study the effect of scale-adaptive self attention (SASA) in Tab. \ref{table:mssa}.
Compared with the vanilla multi-head self attention, SASA achieves +4.0 mAP and +2.2 NDS improvements over the baseline. We further ablate different distance functions and find that $L_2$ distance works best. Besides, we also observe two interesting phenomena in SASA. \textit{First}, different heads learn a different receptive field (see Fig. \ref{fig:mssa_receptive_field}), enabling the model to aggregate multi-scale features in BEV space. \textit{Second}, queries representing larger objects tend to have larger receptive field. In Fig. \ref{fig:sasa_class}, we average the receptive field coefficient $\tau$ over all queries and all heads for each class. As we can see, the receptive field of large objects (such as bus and truck) is larger than small objects (such as pedestrian and traffic cone), demonstrating the effectiveness of our adaptive design.

\begin{figure}[t]
  \centering
  \includegraphics[width=\linewidth]{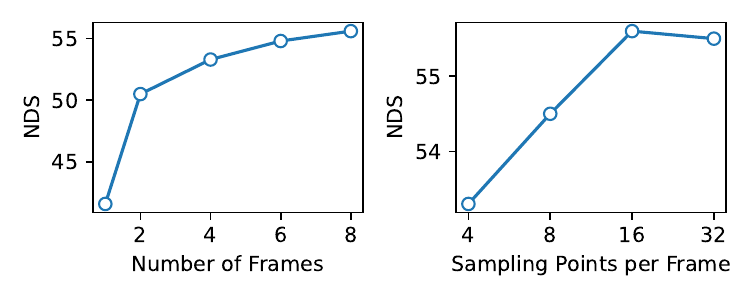}
  \vspace{-20pt}
  \caption{Ablations of adaptive spatio-temporal sampling. The performance continues to increase with the number of frames. For sampling points, we observe that 16 points per frame works best.}
  \label{fig:ablation_sampling}
\end{figure}

\begin{figure*}
  \center
  \captionsetup[subfigure]{labelformat=empty}

  \begin{subfigure}[b]{0.166\linewidth}
      \caption{\texttt{FRONT\_LEFT}}
  \end{subfigure}%
  \hfill
  \begin{subfigure}[b]{0.166\linewidth}
      \caption{\texttt{FRONT}}
  \end{subfigure}%
  \hfill
  \begin{subfigure}[b]{0.166\linewidth}
      \caption{\texttt{FRONT\_RIGHT}}
  \end{subfigure}%
  \hfill
  \begin{subfigure}[b]{0.166\linewidth}
      \caption{\texttt{BACK\_RIGHT}}
  \end{subfigure}%
  \hfill
  \begin{subfigure}[b]{0.166\linewidth}
      \caption{\texttt{BACK}}
  \end{subfigure}%
  \hfill
  \begin{subfigure}[b]{0.166\linewidth}
    \caption{\texttt{BACK\_LEFT}}
  \end{subfigure}%
  \hfill

  \vspace{-1pt}
  \includegraphics[width=\linewidth]{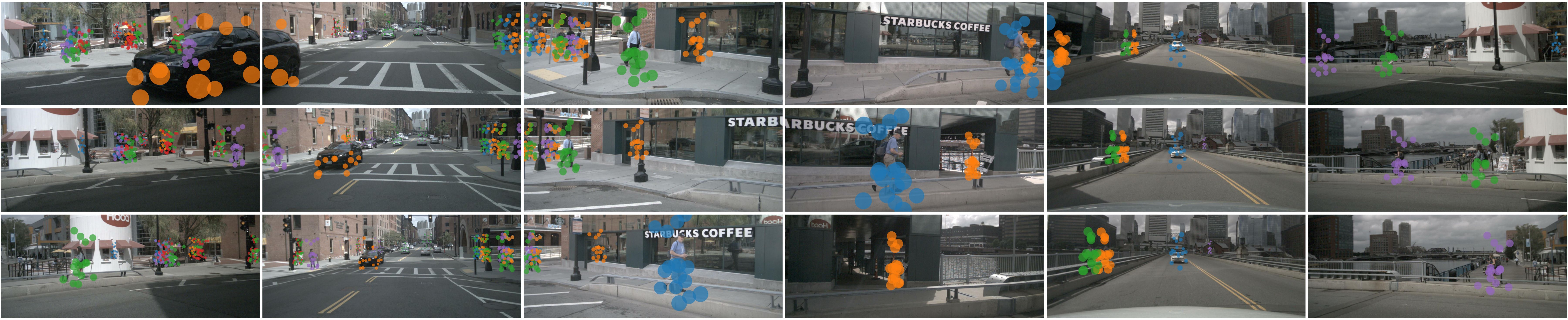}
  \vspace{-15pt}
  \caption{Visualization of adaptive spatio-temporal sampling over three consecutive frames. Different instances are distinguished by colors. Point size indices the depth: larger points are closer to the camera. Our sampling scheme has an adaptive receptive field and is well aligned across different timestamps.}
  \label{fig:viz_nuscenes_sp}
  \vspace{-5pt}
\end{figure*}

\paragraph{Adaptive Spatio-temporal Sampling.} In Fig. \ref{fig:ablation_sampling}, we ablate the number of frames and sampling points. The performance continues to improve as the number of frames increases, proving that SparseBEV can benefit from the long-term history. Here, we use 8 frames for fair comparison with previous methods. As for the number of sampling points, we observe that dispatching 16 points for each frame leads to the best performance. We also provide the visualization of the sampling points from the last stage of the decoder in Fig. \ref{fig:viz_nuscenes_sp}. Our sampling scheme has adaptive regions of interest and achieves good temporal alignments for both static and moving objects across different frames.

\begin{table}[t]
  \centering
  \begin{tabular}{cc|ccc}
    \toprule
    Ego Align & Obj. Align & NDS & mAP & mAVE \\
    \midrule
    - & - & 44.4 & 34.1 & 0.510 \\
    $\surd$ & - & 54.2 & 43.5 & 0.281 \\
    $\surd$ & $\surd$ & \textbf{55.6} & \textbf{45.4} & \textbf{0.243} \\
    \bottomrule
  \end{tabular}
  \caption{Ablations on temporal alignment. Aligning both ego and object motion leads to the best performance.}
  \label{table:temporal_align}
  \vspace{-6pt}
\end{table}

\vspace{-5pt}
\paragraph{Temporal Alignment.} In Tab. \ref{table:temporal_align}, we validate the necessity of temporal alignment in spatio-temporal sampling. In our method, ego motion is aligned with the provided ego pose, while object motion is aligned with a simple constant velocity motion model. As we can see from the table, both of them contribute to the performance.

\vspace{-5pt}
\paragraph{Adaptive Mixing.} In Tab. \ref{table:mixing}, we validate the design of the adaptive mixing.
The top row is a baseline that uses attention weights to aggregate sampled features (as done in DETR3D).
Static mixing and adaptive mixing improve the baseline by 2.7 NDS and 6.5 NDS respectively, demonstrating the necessity of the adaptive design.
Next, we explore different combinations of channel and point mixing. Channel mixing followed by point mixing leads to better performance, proving that the object semantics should be enhanced before point mixing.

\subsection{Limitations and Future Work}

One limitation of SparseBEV is the heavy reliance on ego pose. As we can see from Tab. \ref{table:temporal_align}, the performance drops about 10 NDS without ego-based temporal alignment. However, in the real world, the ego pose provided by IMU may be unreliable and inaccurate, seriously affecting the robustness. Another limitation is that the inference latency increases linearly with the number of frames, since we stack the sampled features along the temporal dimension.

In the future, we will explore a more elegant and concise way of decoupling spatial appearance and temporal motion. We will also try to extend SparseBEV to other 3D perception tasks such as BEV segmentation, occupancy prediction and lane detection.

\begin{table}[t]
  \centering
  \begin{tabular}{c|c|cc}
    \toprule
    Method & Details & NDS & mAP \\
    \midrule
    W/o Mixing & - & 49.1 & 38.6 \\
    Static Mixing & Channel $\rightarrow$ Point & 51.8 & 41.9 \\
    \midrule
    \multirow{4}{*}{Adaptive Mixing} & Channel Only & 53.1 & 42.7 \\
    & Point Only & 53.6 & 43.3 \\
    & Point $\rightarrow$ Channel & 55.4 & 44.6 \\
    & Channel $\rightarrow$ Point & \textbf{55.6} & \textbf{45.4} \\
    \bottomrule
  \end{tabular}
  \caption{Ablations on adaptive mixing. Channel mixing followed by point mixing leads to better performance.}
  \label{table:mixing}
  \vspace{-6pt}
\end{table}

\section{Conclusion}

In this paper, we have proposed a query-based one-stage 3D object detector, named SparseBEV, which can enjoy the benefits of the BEV space without explicitly constructing a dense BEV feature. SparseBEV improves the adaptability of the decoder by three key modules: scale-adaptive self attention, adaptive spatio-temporal sampling and adaptive mixing. We further introduce a dual-branch design to enhance the long-term temporal modeling. Experiments show that SparseBEV achieves the state-of-the-art performance on the dataset of nuScenes for both speed and accuracy. We hope this exciting result will attract more attention to the fully sparse BEV detection paradigm.

\paragraph{Acknowledgements.} This work is supported by National Key R$\&$D Program of China (No. 2022ZD0160900), National Natural Science Foundation of China (No. 62076119, No. 61921006), Fundamental Research Funds for the Central Universities (No. 020214380091, No. 020214380099), and Collaborative Innovation Center of Novel Software Technology and Industrialization. Besides, the authors would like to thank Ziteng Gao, Zhiqi Li and Ruopeng Gao for their help.

{\small
\bibliographystyle{ieee_fullname}
\bibliography{egbib}
}

\begin{figure*}[t]
  \centering
  \includegraphics[width=\linewidth]{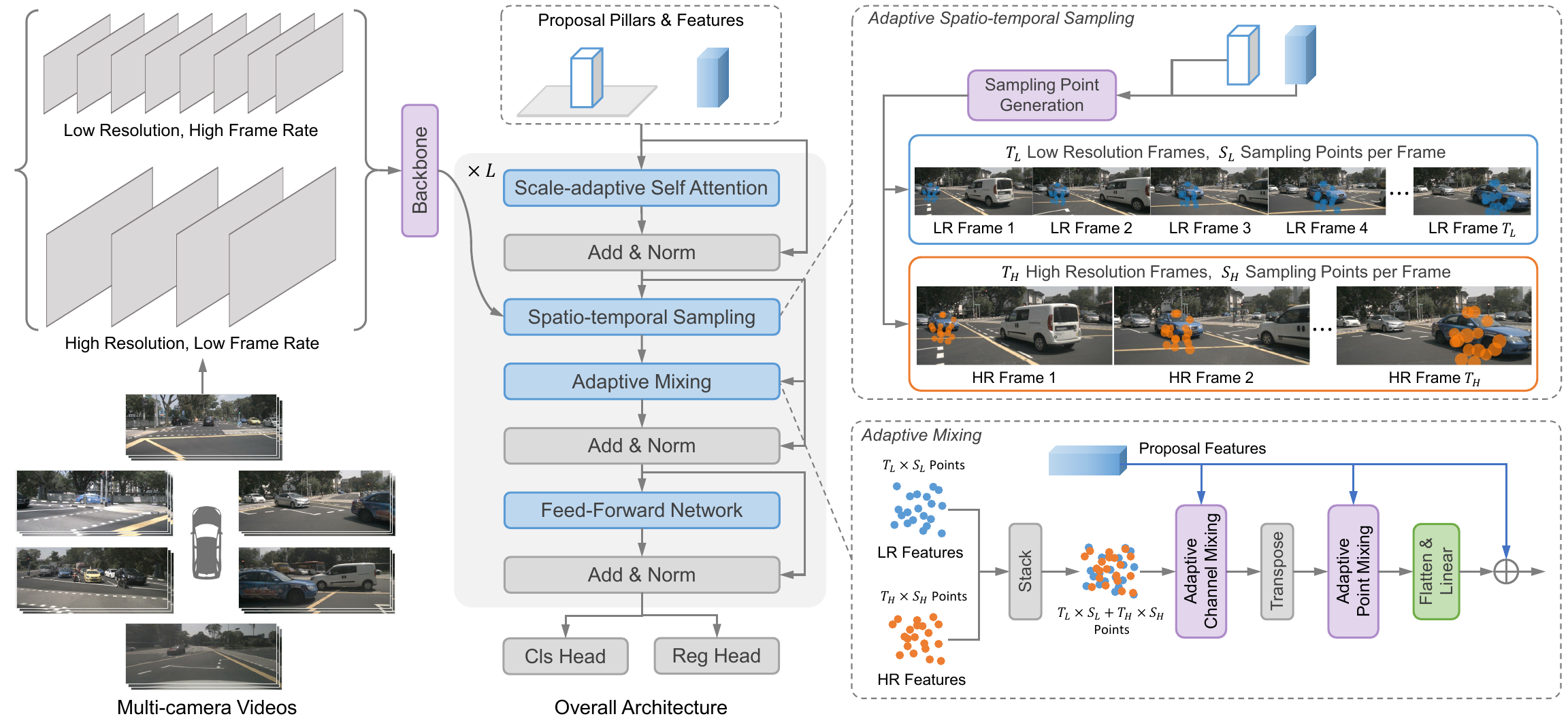}
  \vspace{-10pt}
  \caption{Architecture of dual-branch SparseBEV. The input multi-camera videos are divided into a high-resolution ``slow'' stream and a low-resolution ``fast'' stream.}
  \label{fig:dual_branch_arch}
\end{figure*}

\newpage
~
\newpage

\appendix

\section{Details of Dual-branch SparseBEV}

In this section, we provide detailed explanations and ablations on the dual branch design. As shown in Fig. \ref{fig:dual_branch_arch}, the input multi-camera videos are divided into a high-resolution ``slow'' stream and a low-resolution ``fast'' stream. Sampling points are projected to the two streams respectively and the sampled features are stacked before adaptive mixing. Experiments are conducted with a V2-99 backbone pretrained by FCOS3D \cite{fcos3d} on the training split of nuScenes. \footnote{Note that the experiment setting used here is different from that in the main paper, since the experiments are conducted before the submission of ICCV 2023. After submission, we further improve our implementation to refesh our results. The conclusion is consistent between these different implmentations.}

\begin{figure}[t]
  \centering
  \includegraphics[width=\linewidth]{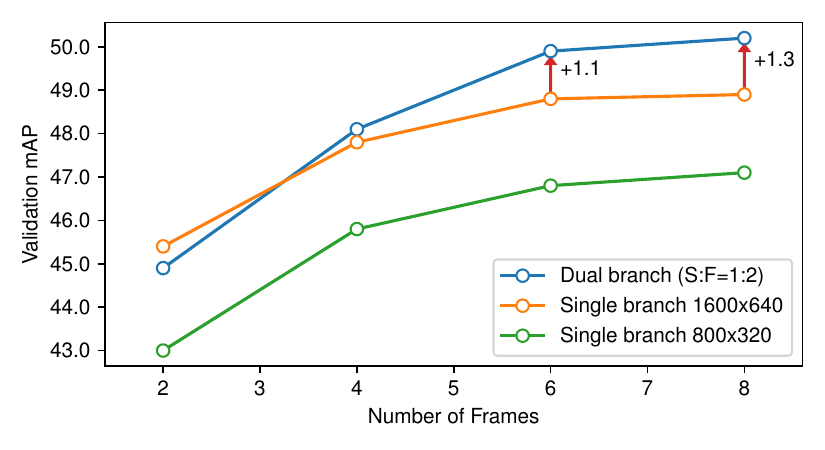}
  \vspace{-20pt}
  \caption{Comparison between single-branch and dual-branch under different settings. Dual branch design brings more gain as the number of frames increases.}
  \label{fig:dual_branch}
\end{figure}

In Fig. \ref{fig:dual_branch}, we compare our dual branch design with single branch baselines. If we use a single branch of 1600 $\times$ 640 (orange curve) resolution, adding more frames does not provide as much benefit as it does at 800 $\times$ 320 resolution (green curve). By using dual branch of 1600 $\times$ 640 and 640 $\times$ 256 resolution with 1:2 ratio, we decouple spatial appearance and temporal motion, unlocking better performance. As we can see from the blue curve, the longer the frame sequence, the more gain the dual branch design brings.

In Tab. \ref{table:dual_branch}, we provide detailed quanlitative results. Under the setting of 8 frames ($\sim$ 4 seconds), our dual branch design with only \textit{two} high resolution (HR) frames surpasses the baseline with \textit{eight} HR frames. By increasing the number of HR frames to 4, we further improve the performance by 0.8 mAP and 0.5 NDS. Moreover, increasing the resolution of the LR frames does not bring any improvement, which clearly demonstrates that appearance detail and temporal motion are decoupled to different branches.

\begin{table}[t]
  \centering
  \begin{tabular}{l|c|cc}
    \toprule
    Method & Setting & mAP & NDS \\
    \midrule
    Single branch & 8f $\times$ 1600 & 48.9 & 57.3 \\
    Dual branch & 2f $\times$ 1600 + 8f $\times$ 640 & 49.4 & 57.9 \\
    Dual branch & 4f $\times$ 1600 + 8f $\times$ 640 & \textbf{50.2} & \textbf{58.4} \\
    Dual branch & 4f $\times$ 1600 + 8f $\times$ 800 & 50.1 & 58.0 \\
    \bottomrule
  \end{tabular}
  \caption{Ablations on the dual branch design. Nf $\times$ M indices the number of frames is $N$ and the longer side of the image has $M$ pixels. For example, ``8f $\times$ 640'' denotes 8 frames with $640 \times 256$ resolution.}
  \label{table:dual_branch}
\end{table}

Since the dual-branch design also enlarges the receptive field (smaller resolution provides larger receptive field) which may improve performance, we further analyse where the improvement comes from in Tab. \ref{table:dual_branch_c6}. The first row is our baseline which takes 8 frames with a single branch of $1600 \times 640$ as input. We first try to increase the receptive field by adding an extra $C_6$ feature map (Row 2), and observe that the performance is slightly improved. This demonstrates that a larger receptive field is required for high-resolution and long-term inputs. However, the spatial appearance and temporal motion is still coupling, limiting the performance. By using dual branches of 1600 $\times$ 640 and 640 $\times$ 256 with 1:2 ratio (Row 3), we decouple spatial appearance and temporal motion, leading to better performance. Moreover, the training cost is also reduced by 1/3. This experiment demonstrates that we not only need larger receptive fields, but also decouple spatial appearance and temporal motion.

\begin{table}[t]
  \centering
  \begin{tabular}{l|c|c|c}
     \toprule
     Method & Feature Maps & Train. Cost & mAP \\
     \midrule
     Single branch & $C_2, C_3, C_4, C_5$ & 2d 17h & 48.9 \\
     Single branch & $C_2, C_3, C_4, C_5, C_6$ & 2d 18h & 49.3 \\
     Dual branch & $C_2, C_3, C_4, C_5$ & \textbf{1d 19h} & \textbf{50.2} \\
     \bottomrule
  \end{tabular}
  \caption{Detailed analyses on the dual-branch design. For single branch baselines, simply adding an extra $C_6$ feature map has limited effect. In contrast, our dual branch design can boost the performance significantly.}
  \label{table:dual_branch_c6}
\end{table}  

\section{Study on Scale-adaptive Self Attention}

\begin{table}[t]
  \centering
  \begin{tabular}{c|c|cc}
    \toprule
    Self Attention & Distance Function & NDS & mAP \\
    \midrule
    SASA-beta & $\tau D$ & 55.2 & 44.8 \\
    SASA      & $\tau D$ & \textbf{55.6} & \textbf{45.4} \\
    \bottomrule
  \end{tabular}
  \caption{Compared with SASA-beta, SASA not only has the ability of multi-scale feature aggregation, but generates adaptive receptive field for each query as well.}
  \label{table:sasa_beta}
\end{table}

In this section, we'll talk about how we came up with scale-adaptive self attention (SASA).
In the main paper, the receptive field coefficient $\tau$ is specific to each head and adaptive to each query. In the development of SASA, there is an intermediate version (dubbed SASA-beta for convenience): the $\tau$ for each head is simply a learnable parameter shared by all queries. 

\begin{figure}[t]
  \centering
  \subfloat{\includegraphics[width=0.5\linewidth]{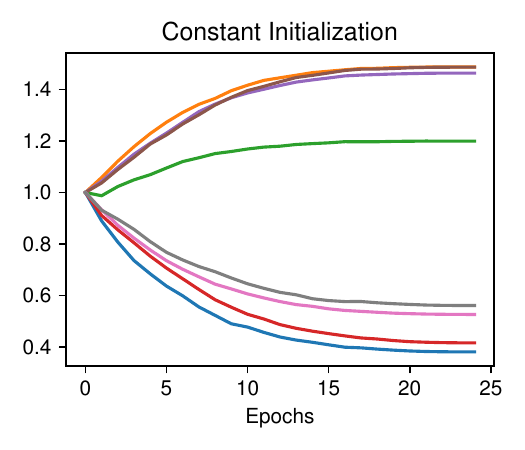}}%
  \hfill
  \subfloat{\includegraphics[width=0.5\linewidth]{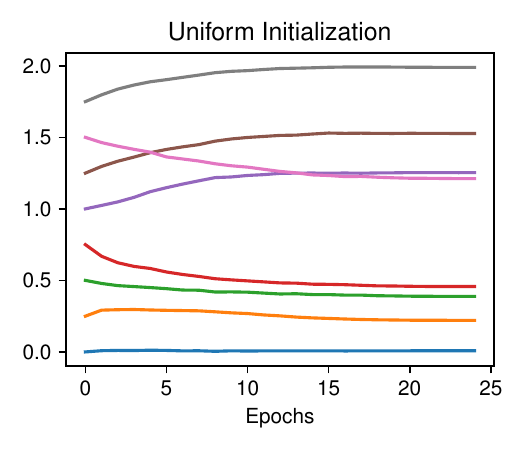}}%
  \vspace{-5pt}
  \caption{The change of $\tau$ of each head in SASA-beta during training. Regardless of the initialization, each head learns a different $\tau$, enabling local and multi-scale feature aggregation.}
  \label{fig:sasa_tau_training}
\end{figure}

In Fig. \ref{fig:sasa_tau_training}, we take a closer look at how $\tau$ changes with training. We surprisingly find that regardless of the initialization, each head learns a different $\tau$ from the others and all of them are distributed in range $[0, 2]$, enabling the network to aggregate local and multi-scale features from multiple heads.

Next, we improve SASA-beta by generating the $\tau$ adaptively from the query, which corresponds to the version in the main paper. Compared with SASA-beta, SASA not only has the ability of multi-scale feature aggregation, but generates adaptive receptive field for each query as well. The quanlitative comparison between SASA-beta and SASA is shown in Tab. \ref{table:sasa_beta}.

\section{More Visualizations}

\begin{figure*}
  \centering

  \subfloat[Sample 0005, stage 1]{\includegraphics[width=\linewidth]{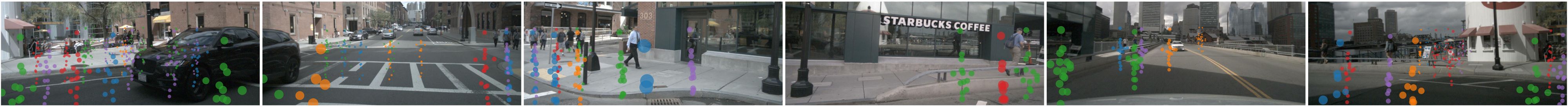}}%
  \vspace{5pt}
  \subfloat[Sample 0005, stage 2]{\includegraphics[width=\linewidth]{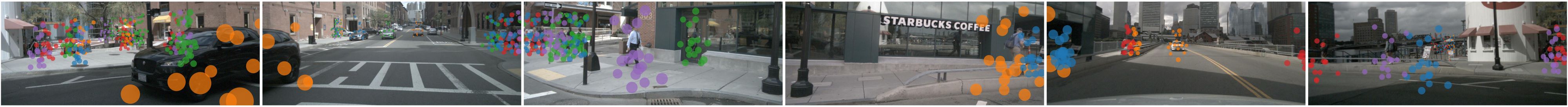}}%
  \vspace{5pt}
  \subfloat[Sample 0005, stage 3]{\includegraphics[width=\linewidth]{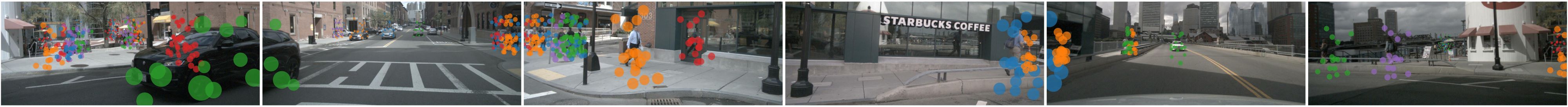}}%
  \vspace{15pt}
  \subfloat[Sample 0028, stage 1]{\includegraphics[width=\linewidth]{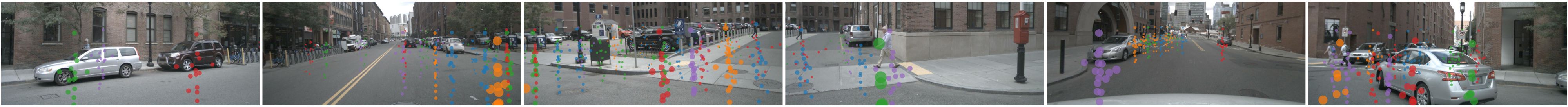}}%
  \vspace{5pt}
  \subfloat[Sample 0028, stage 2]{\includegraphics[width=\linewidth]{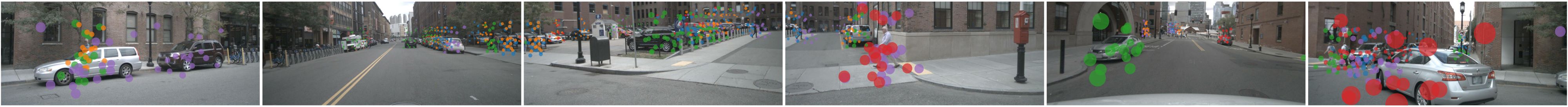}}%
  \vspace{5pt}
  \subfloat[Sample 0028, stage 3]{\includegraphics[width=\linewidth]{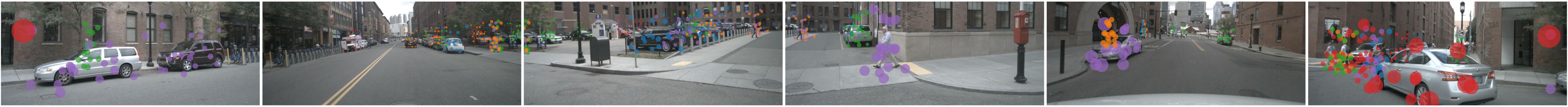}}%
  
  \caption{Visualized sampling points from different stages. Different instances are distinguished by colors.}
  \vspace{-5pt}
  \label{fig:sp_stage}
\end{figure*}

In Fig. \ref{fig:sp_stage}, we provide more visualizations of the sampling points from different stages. In the initial stage, the sampling points have the shape of pillars. In later stages, they are refined to cover objects with different sizes.

\end{document}